# Predicting litigation likelihood and time to litigation for patents

Papis Wongchaisuwat[1]; Diego Klabjan[1]; John O. McGinnis[2]

[1]Department of Industrial Engineering and Management Sciences,

Northwestern University, Evanston, IL

[2]Northwestern University School of Law, Northwestern University, Chicago, IL

**Abstract**

Patent lawsuits are costly and time-consuming. An ability to forecast a patent litigation and time to litigation allows companies to better allocate budget and time in managing their patent portfolios. We develop predictive models for estimating the likelihood of litigation for patents and the expected time to litigation based on both textual and non-textual features. Our work focuses on improving the state-of-the-art by relying on a different set of features and employing more sophisticated algorithms with more realistic data. The rate of patent litigations is very low, which consequently makes the problem difficult. The initial model for predicting the likelihood is further modified to capture a time-to-litigation perspective.

## 1. Introduction

According to the U.S. Patent and Trademark Office (USPTO), large numbers of patents are issued yearly. The number of patents granted annually has trended upward in the past decade. More importantly, the U.S. Government Accountability Office reported an increasing trend in the number of patent infringement lawsuits filed in the U.S. district courts over the past decade[1]. Figure 1 presents the total number of granted patents from 2000 to 2011, as reported by USPTO's Patent technology monitoring team[2] compared with a slight fluctuation in the number of patent lawsuits from 2000 to 2010. This is a 31% increase in the number of lawsuits from 2010 to 2011.

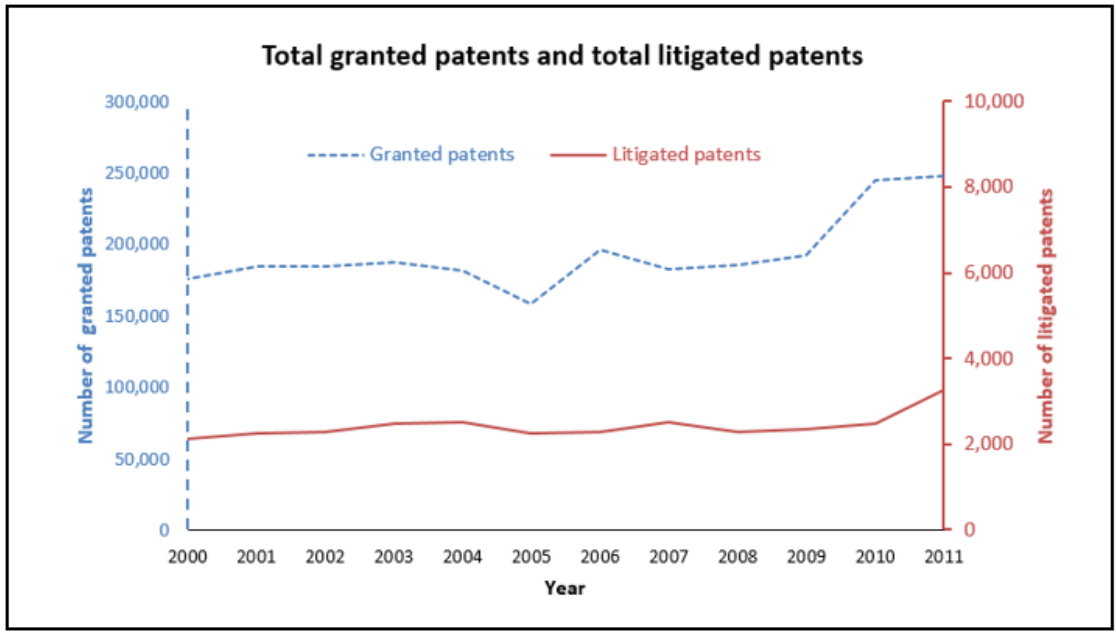

Figure 1: Total granted patents and patent litigations each year[1, 2]

A litigation is commonly associated with high cost of patent lawsuits and time-consuming legal processes. Patents are means to protect intellectual property and they establish inventions. Some large companies including Canon, Google, IBM, Microsoft and Samsung have been obtaining and accumulating a large number of patents each year, especially in recent years[3].

---

[1] USPTO's Patent technology monitoring team, U.S. patent statistics chart calendar years 1963-2013.
http://www.uspto.gov/web/offices/ac/ido/oeip/taf/us_stat.htm
[2] U.S. Government accountability office, intellectual property: assessing factors that affect patent infringement litigation could help improve patent quality; 2013
http://www.uspto.gov/sites/default/files/aia_implementation/GAO-12-465_Final_Report_on_Patent_Litigation.pdf
[3] Center for global innovation/patent matrics, Global patent quality statistics & investment analysis
http://www.bustpatents.com/statistics.htm

Companies typically allocate significant resources to prevent others from infringing upon their patents. A budget for monitoring possible infringements can be better allocated if the companies are able to predict which patent and when it is likely to be litigated. Patent trolls who accumulate third party patents, instead of investing a large amount of money to collect a portfolio of patents, could improve their portfolio selection by having the ability to accurately indicate whether a patent has a high chance of being contested. A patent troll could also take advantage of time-to-litigation predictions by better forecasting an exact time to purchase a patent. This work develops predictive models to differentiate between likely-to-be litigated and not-to-be-litigated patents ahead of time as well as to predict when the litigation is going to take place. Hence, the models developed in our work may help companies achieve a more realistic budget allocation plan or improve patent portfolios of patent trolls.

We develop a combination of clustering and classification models to predict which patent is likely to be litigated based on multiple features including textual types, which are extracted directly from the claim section in a patent, as well as non-textual types, which are obtained from other relevant information. The data set used in our work is highly imbalanced between litigated and non-litigated cases hence a re-sampling method is implemented prior to fitting the classification models. In the second part of the paper, we aim to forecast the time to litigation of disputed patents. The models are tested on different data sets including patents associated with three keywords: "Wireless Network," "Advertising" and "Telecommunication." These keywords are related to the technology industry, and, unsurprisingly, technology companies hold large portfolios of patents. These companies commonly compete with each other, which can lead to a controversy regarding the ownership of an invention. Consequently, technology companies have a higher chance of being involved in patent lawsuits; we have therefore specifically selected technology-related keywords to test the models. We develop the models based on three options of auxiliary data sets augmenting other features with different financial information (obtained from the U.S. Securities and Exchange Commission - SEC data). The three data options include the model without the SEC data, with the SEC data by eliminating records without the SEC information, and with the SEC data under assumptions regarding the default values for missing observations. In order to measure the performance of the model, we use common metrics including the precision and recall, the F1-score, and the confusion matrix [1]. The overall F1-score is used to evaluate each class in the time-to-litigation model since this model is multiclass.

Given a test patent to be predicted, our models use a clustering approach combined with a heuristic technique as well as an ensemble classification method to predict the litigation likelihood and the time to litigation. In other words, the models specify whether the test patent belongs to the "litigate" class in the litigation model. For the time-to-litigation model, we categorize all cases into different year groups based on their time to litigation information. We apply a similar algorithm as the litigation model to each year group and finally re-adjust the predicted classes by taking the hierarchical relationship among year groups into consideration. For example, if a patent is predicted to be litigated before year 4, then it is also predicted to be litigated before year 5. The K-means algorithm is implemented as the baseline clustering approach for both models. The distance between the test case and convex hulls of the clusters are computed to determine which class it belongs to. The heuristic technique as well as the ensemble classification method between Support Vector Machine (SVM) and random forest are further used to re-estimate the class of the test case when the designation is too ambiguous from clustering.

Based on our study, the "Wireless Network" keyword with SEC data assuming no default value yields the highest F1-score of 0.19 for the litigation model. Among the three keywords,

"Wireless Network" tends to give the highest F1-score, followed by "Telecommunication" and "Advertising" performs the worst. Using SEC information without a default value yields the best result while assuming a default value performs better than excluding SEC data. For the time-to-litigation model, "Wireless Network" generally yields a higher F1-score for most classes and data options than the other two keywords. We observe that enhancing the classification method with clustering does not improve the time-to-litigation model's performance compared to utilizing only the classification model.

The features, the models, and the data sets used to test the models distinguish our work from others. Specifically, our work has three main contributions. First, we explore other informative features that have not been studied in prior works. We include number of referenced patents that were litigated, the second layer of references as well as PageRank score features in the models to capture deeper knowledge of referencing. We also introduce financial information of the patent's assignee into the models by including SEC data features. Second, we use a combination of clustering methods adapted from anomaly detection models enhanced with a standard classification approach in the litigation model. We cluster litigated and non-litigated models and use convex hulls of clusters which has not yet been done in the past. Also, no work has previously been proposed in predicting the time to litigation for disputed patents. Finally, we test the performance of all models with the testing data that reflects an actual rate of litigations, i.e. the severe imbalance of classes.

In Section 2, we describe our models thoroughly including features used. Further description regarding data collection is provided in Section 3. The results of our models and further discussions are reported in Section 4. We conclude the introduction with a literature review.

**Relevant work**

A patent is an intellectual right granted to an inventor in order to prevent others from using the same invention. Each year a large number of patents are granted by USPTO, and some of them are infringed. This inevitably leads to litigations. Chien [2] studied various factors influencing the likelihood of a patent litigation. Instead of focusing on intrinsic factors which are embedded within a patent, acquired characteristics developed after the patent has been issued were specifically analyzed. These acquired features include changes in patent ownership, continued investment in the patent, securitization of the patent, and citations to the patent. In the experimental data set, each randomly selected litigated patent was combined with three additional patents issued in the same year and assigned to the same class. Consequently, the reported resulting rate of litigation is significantly higher than the actual rate of approximately 1 to 2 percent. The results indicated that patents having a higher chance of getting disputed can be distinguished in advance from those being less likely to be involved in a dispute. Chien reported the predictability of the model by comparing the number of patents predicted to be litigated versus the number of actually litigated patents. Although the performance reported in [2] is better than ours, the litigation ratio assumed in [2] is much higher. The author's work undoubtedly made a significant contribution to the field of predicting patent litigations. Petherbridge [3] pinpointed the limitation of Chien's model in terms of accuracy and practicability including a high false positive rate. Comparing with our work, [3] provided an explanation illustrating a relatively rare event of the patent litigation problem while we tested the models with actual data. Even though our model shows the similar issue as Chien's work of a high misclassification potential, it can be calibrated to reduce the false positive rate under an acceptable false negative rate. Kesan et al. [4] provided follow-up research to Chien's work where multiple flaws and possible improvements were discussed including the data collection, the features, and the normative conclusions. In summary, these works discussed limitations of Chien's work in both methodological and usage perspectives. They provided some possible solutions without implementing the actual models. We address some of these flaws. In particular, our models use actual data covering a larger time span as pinpointed by [3] and [4]. We introduce different informative features that have not yet been considered in prior works. These features can be gathered when the patents are issued, which addresses the issue stated in [4] that some features used by Chien such as whether the patent was reassigned and whether the patent was in reexamination cannot be obtained at the time of the prediction. It is likely that the patent is often reassigned only after realizing the litigation's decision [4].

Moreover, empirical factors determining which patent is more likely to be litigated were studied in [5] and [6]. According to these studies, both information on patent lawsuits and patent documents were collected and combined to identify how the characteristics of patents affect the likelihood of litigation. Multiple hypotheses were statistically tested. The number of claims, the technology-based classification system of the patent, citations, country and type of ownership, and the size of the patent portfolio were considered as factors in these papers. An empirical analysis of determinants of a patent litigation in a German court was studied in [7]. Similar results based on the US legal system were obtained, except that there was no difference between the chance of facing litigation among individual patent owners and companies in the German system. Unlike our work, which implements machine-learning algorithms as predictive models, [5], [6] and [7] mainly focused on studying determinants of patent lawsuits rather than their prediction.

Another interesting line of research is predicting the outcome of a patent litigation. Cowart et al. [8] implemented a logistic regression model and a classification tree to predict a legal decision making process. In the study, both algorithms provided a similar overall prediction rate of 78 percent. Kashima et al. [9] developed a model predicting patent quality by measuring the stability of a patent, which is indicated by the possibility of patent surviving in court. The work employed a machine learning technique to predict the outcome of the IP High Court in Japan. The results showed the performance of 0.65 Area Under the Curve. The model proposed by Kashima et al. relied on both textual and non-textual features. It focused on predicting a patent quality or the court outcome while ours aim to predict the litigation likelihood. The evaluation measures are also different as AUC used in [9] is not suitable for a highly imbalanced data set like ours.

Our problem is also related to the anomaly detection task where the number of anomalous items is relatively small compared to the whole data set. In our case, a litigation (anomaly) is a rare event. Three main categories of anomaly detection techniques including supervised, semi-supervised and unsupervised are reviewed in [10]. With availability of labeled data, a predictive model for normal and anomalous classes is typically constructed. We instead use the clustering technique, a common unsupervised approach, to enhance the performance of the classification model due to the lack of a large number of historical records, i.e., patents. To the best of our knowledge, we also contribute in this space since we cluster both classes and combine clustering with classification.

## 2. Method

Our goals are to predict which patent is likely to be disputed and when it would occur. We construct two main models, including the litigation likelihood model and the time-to-litigation model. The clustering approach combined with a variation of the nearest convex hull classification [11] is improved with the ensemble classification model. To address the issue of imbalanced data sets in training the classification models, we re-sample the data set. Relevant features are needed in order to train the models. We use the same set of features in both litigated and time-to-litigation models, which are discussed in detail next.

2.1. Features

Features used in the models are divided into two distinct groups: textual and non-textual features.

2.1.1     Textual features: We rely on the assumption that words occurring in a patent contain significant information in determining the litigation likelihood. Each patent consists of detailed information including the claim section from where we extract textual features. A document-term matrix is constructed based on all claims mentioned in the patent by incorporating unigram (one word), bigram (a pair of words) and trigram (a triple of words) features. The values in the matrix correspond to the term frequency-inverse document frequency (tf-idf) factor. The tf-idf value for a particular word increases if it appears frequently in the document but decreases when it relatively appears often in all documents. This idea takes into account the fact that some common words appearing frequently in general should not be important. The generated matrix is large and thus we select 30 textual features with the highest information gain to be the final set of textual features.

2.1.2     Non-Textual features: In addition to the knowledge embedded in the claim section of a patent, other relevant information should be considered. These features which are extracted from the patent document and SEC websites are listed in Table 1. Financial data is discounted to the current time and categorized into groups. All features 1-9 are numerical values while features 10-12 are categorical values.

Table 1: List of features included in the model and their source of information

| Non-textual features | Sources |
|---|---|
| **General information about the patent** | |
| 1. Number of inventors | The patent document |
| 2. Number of claims | The patent document |
| 3. Number of words in claims | The patent document |
| 4. Number of foreign references | The patent document |
| 5. Number of backward references | The reference network constructed from patent documents |
| 6. Number of $2^{nd}$ layer of backward references | The reference network constructed from patent documents |
| 7. Number of litigated backward references | The reference network constructed from patent documents |
| 8. Number of $2^{nd}$ layer of litigated backward references | The reference network constructed from patent documents |
| 9. Average of PageRank score of backward references | The reference network constructed from patent documents |
| **Financial data of the patent's assignee** | |
| 10. Revenue | The SEC website |
| 11. Earnings per share | The SEC website |
| 12. Market share price | The SEC website |

The $2^{nd}$ layer of reference conveys more insights on how a patent relates to others, i.e. indicating patents within a similar area of interest. The PageRank score is developed by Google to rank websites in the search engine results, which is used to measure the significance of each web page. Applying this idea to our framework, we implement the PageRank score with the reference network constructed from patent documents. In this network, each node corresponds to a patent and there is an edge if and only if there is a reference relationship between the two patents. The PageRank score of each node is a weighted average of PageRank scores of all connected nodes.

The weight is assigned based on the significance of each node, i.e. a reciprocal of the number of outgoing edges. The basic idea behind this concept is that the more important a patent is, the more likely it receives links referenced from other patents.

The likelihood of litigation for patents is potentially related to the business aspects of companies owning the patents. Three features collected from the SEC website including revenue, earnings per share and market share price represent the financial information of the patent's assignee.

2.2. The litigation model

Our first attempt to predict the litigation likelihood is utilizing the standard classification approach where the supervised learning algorithms of SVM, decision tree, boosted tree, random forest, as well as ensemble methods among these algorithms are experimented with. We call this approach *pure classification*. The ensemble model between SVM and random performs best. The classification method performs satisfactorily but not as well as the following algorithm named *the cluster with ensemble method*.

We first cluster litigated and non-litigated cases in the training set by using the K-means algorithm as depicted in Figure 2. In what follows, we treat 1/0 as infinity applied to all ratio computations. The flow diagram of the cluster with ensemble algorithm is shown in Figure 3. In scoring, first, the distance between a test case and the convex hull constructed with all members in each cluster is computed. The ratio of the distance between the test case to the closest litigated and the closest non-litigated cluster, named convex hull distance ratio, is computed. The test case is initially assigned to be litigated if this ratio is smaller than some hyper parameter A, and non-litigated otherwise. We next construct a ball centered at the test case with radius z which is a fraction of hyper parameter X and the distance r between the test case and the closest point of the class to which the test case was previously assigned to. We next compute the ratio between the number of litigated and non-litigated cases falling inside the ball, named litigated fraction ratio. If the ratio is lower than some hyper parameter B, the initial litigated label is assigned to be the non-litigated class while the initial non-litigated label remains the same. The ensemble classification model between SVM and random forest is applied to re-adjust the label when the litigated fraction ratio is greater than hyper parameter B.

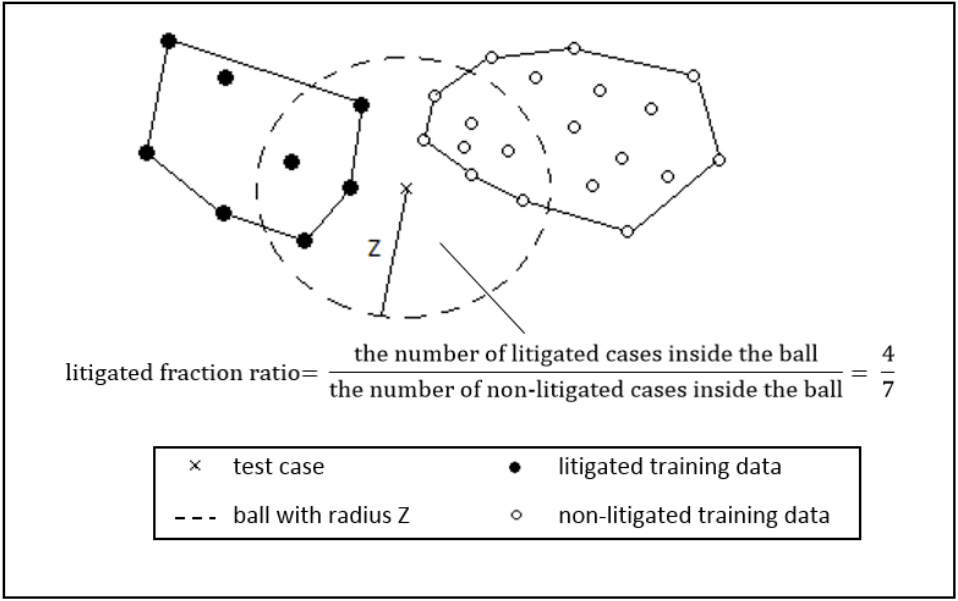

Figure 2: The cluster with ensemble approach

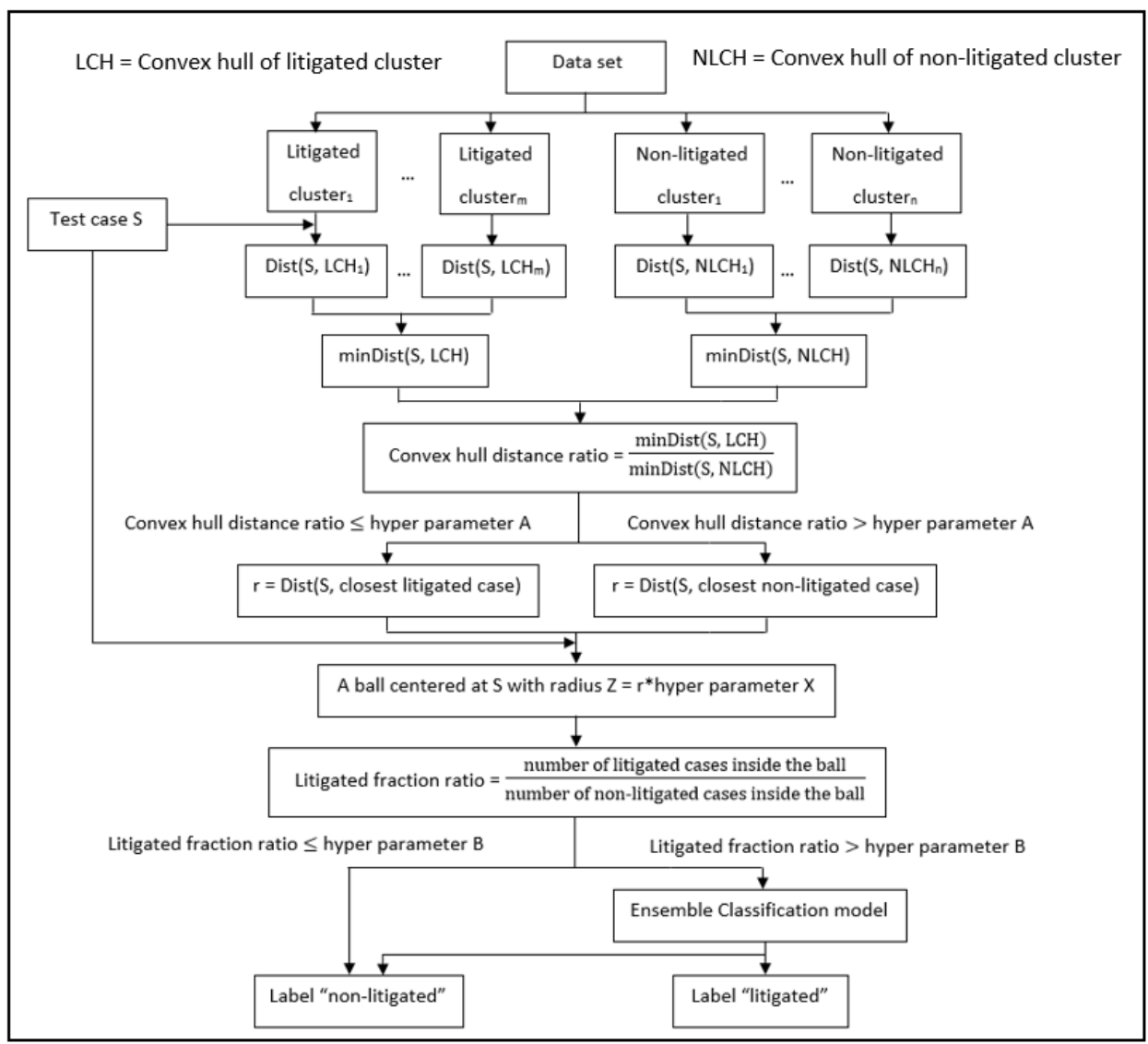

Figure 3: Flow diagram of the cluster with ensemble algorithm

2.3. The time-to-litigation model

The goal of this model is to predict the time to litigation for each disputed case. After collecting the number of years between a patent's issue date and its first litigation date, we categorize all cases into different groups including litigation before 14 years, 7 years, 4 years and 1 year after the issue date of the patent. We set cut-off points between groups by considering big differences in the histogram of time to litigation information in the data set. We use the time-to-litigation groups as the label to fit the models. Then, the final adjustment of a predicted class is implemented by considering that if a patent is predicted to be litigated by year 1, it definitely has to be litigated in later years (by year 4 or 7 or 14). The hierarchical tree indicating time to litigation of the model is illustrated in Figure 4.

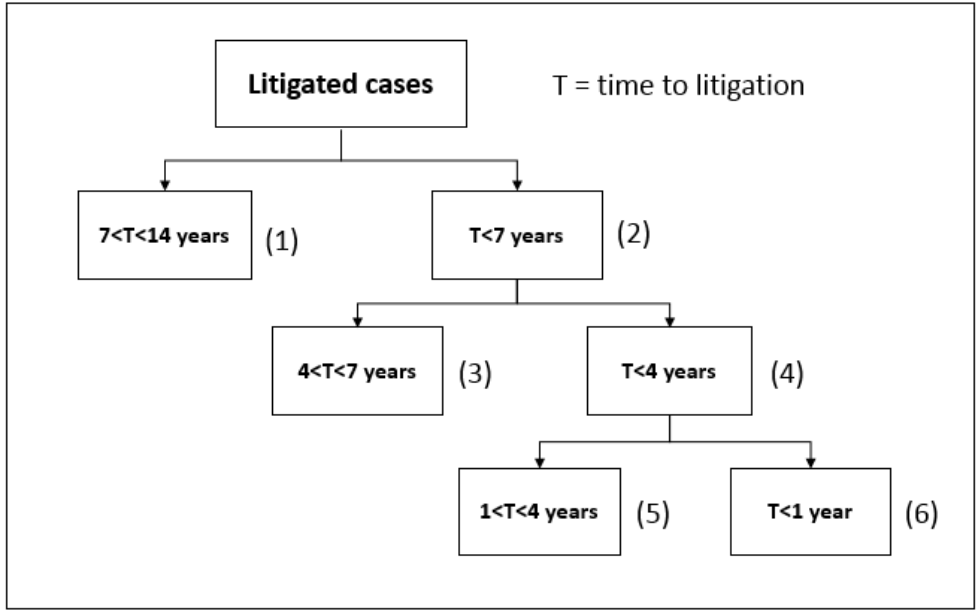

Figure 4: A hierarchical tree for the time-to-litigation model

In addition to fitting a model for each class independently, we also attempt to simultaneously fit a model for all classes while taking the hierarchy of classes into consideration. For the training part, we first cluster the final leaf node in the hierarchical tree (T<1 year) using the K-means algorithm and further expand the clusters with other classes in the hierarchical tree as depicted in Figure 5. We assign each case from the 1<T<4 years class to the closest T<1 cluster depending on the distance to the convex hull of the clusters. We repeat the process for the remaining classes until we achieve 4 layers of classes.

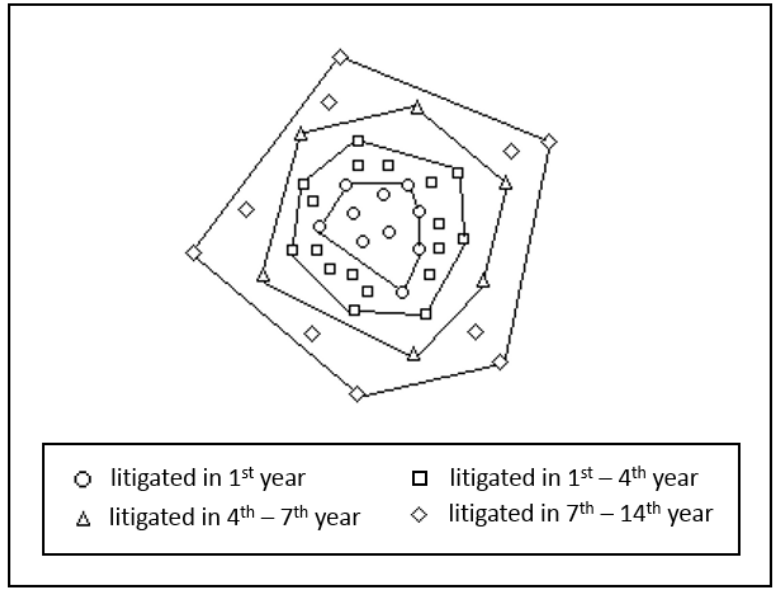

Figure 5: Layers of convex hulls of litigated cases

The time-to-litigation class is assigned based on which layer of the convex hulls each test case falls into. The test case is labeled with the closest convex hull it falls inside. For instance, the test case is classified as the T<4 year class if it falls inside the convex hull of the T<4 year class and outside the convex hull of the T<1 year class. If the test case falls outside the convex hull of T<14 years, it is labeled as T<14 years.

2.4. Re-sampling

The litigation rate is low with a value of 1 to 2 percent among all granted patents. With such an imbalanced data set, it is challenging for an algorithm to perform well. In order to enhance the predictive power of the classification model, the re-sampling technique is used to reduce the unbalancing level in the original data. Specifically, Synthetic Minority Over-sampling Technique (SMOTE) [12], which oversamples the minority class and undersamples the majority class is employed. For the litigation model, the minority and the majority classes used in SMOTE are litigated and non-litigated cases, respectively. We implement SMOTE technique for each label separately for the time-to-litigation model. For example, the minority class for the label "litigated by year 1" (node 6 in Figure 4) is the litigated cases which were only disputed by year 1, while the majority class includes other litigated cases which were disputed after year 1.

2.5. Model calibration and evaluation

After testing the fitted models, we achieve the predicted label directly from the clustering approach and the predicted probability belonging to each class from the classification model. To evaluate the litigation model's performance, traditional metrics including the confusion matrix, precision and recall as well as the F1-score are computed. The F1-score which is the geometric mean between the precision and recall values is used to compare different models. For the time-to-litigation model, the F1-score is computed for each class including nodes 2, 4 and 6 in Figure 4. Among the different machine learning algorithms, SVM and random forest provide satisfactory performance. We consequently implement an ensemble method between these two algorithms by varying different weights given to each algorithm. The ensemble method is further used in the clustering approach which is also calibrated with different hyper parameters to maximize the model's performance. We ran all experiments with multiple replications of 10-fold cross validation to ensure consistency of the results.

**3. Data Collection**

Patent documents were collected based on three keywords: "Wireless Network," "Advertising" and "Telecommunication." We specifically chose these keywords because of their relatively high rate of litigation and many patents. Textual features and the number of inventors, number of claims, and referencing features are gathered directly from a patent. Financial data is collected from the SEC website which requires public companies to file periodic reports (i.e. quarterly and annual reports). The three features revenue, earnings per share and market share price capturing the financial situation of a company were extracted from annual financial reports (10-K for US companies and 20-F for foreign companies).

As this data is very limited due to incomplete information of private or small companies who are not obligated to report to SEC, we fit the models with three data options differing in the SEC data features: the model without the SEC data, with the SEC data, and with the SEC data after assuming a default value for missing cases. In the model with the SEC data, records without the SEC data are omitted before training the model. The default value is assumed to be a reasonable value in practice for each keyword separately. Finally, we collect litigation data from LexMachina[4] to label each instance in our data set. Table 2 illustrates a total number of collected patents, the counts as well as the litigation rate associated with each keyword.

Table 2: Summary of the data set

| Keywords | Without SEC data | | | | With SEC data | | | |
|---|---|---|---|---|---|---|---|---|
| | Litigated cases | Non-litigated cases | Total cases | Litigation rate | Litigated cases | Non-litigated cases | Total cases | Litigation rate |
| Wireless Network | 509 | 26,154 | 26,663 | 1.91% | 156 | 7,511 | 7,667 | 2.03% |
| Advertising | 759 | 19,833 | 20,592 | 3.69% | 134 | 6,068 | 6,202 | 2.16% |
| Telecommunication | 646 | 51,654 | 52,300 | 1.24% | 148 | 9,093 | 9,241 | 1.60% |

**4. Results**

We calibrated the hyper parameters in order to obtain the best performance. SMOTE, the sampling technique, requires two hyper parameters: how many extra minority-class cases are generated, and how many extra majority-class cases are selected for each generated minority case. In our model,

---
[4] Lex Machina. https://lexmachina.com/; Accessed in 2014

5 and 1 are chosen, respectively. For example, assuming that the original data contains 6,500 and 140 records of the majority and minority classes, respectively, the number of minority class records is adjusted to be 840 (140 original cases plus 140*5=700) while the number of majority class records is adjusted to 7,200 (6,500 original plus 700*1=700). For the SVM classification model, a radial kernel with gamma of 0.001 and regularization value of 0.1 are selected. Weighting 0.3 for SVM and 0.7 for random forest gives the best ensemble model. The cut-off probability of 0.3 is chosen. For clustering part, hyper parameter X, A, and B defined in Section 2.2 are set to be 3.5, 1.3, and 0.015, respectively. This set of hyper parameters are applied to both the litigation and the time-to-litigation models.

Information gain is implemented for selecting the most significant features. Features related to reference knowledge are influential factors to the model to indicate the litigation likelihood and time to litigation including the first and second layer of references as well as the litigated references. With SEC data included in the model, features containing financial information also indicate high information gain. The significant non-textual features with their corresponding information gain value for the "Wireless Network" keyword are listed in Table 3. Note that the higher the information gain, the more significant the feature is.

Table 3: Features with their corresponding information gain

| Features | Information gain |
|---|---|
| Revenue | 0.0052 |
| Earnings per share | 0.0035 |
| Litigated backward references | 0.0032 |
| Backward references | 0.0027 |
| 2$^{nd}$ layer of backward references | 0.0019 |

Figure 6 illustrates the top 30 textual features for the "Wireless Network" keyword. Unsurprisingly, common words generally related to the technology industry including "device," "system," "network," "server," "monitoring," "interface," "wireless" are contained in the top 30 features. Moreover, various words occurring in this list are relevant to communication systems, which are closely related to the "Wireless Network" keyword. These specific words include "video," "telephone," "remote," "audio" and they occur more often in litigated than in non-litigated cases. The proportion of litigated cases containing the specific words related to "Wireless Network" and the proportion of litigated cases containing the common words as defined in the beginning of the paragraph relevant to the technology industry are approximately 75 and 30 percent higher than that of non-litigated cases, respectively.

| "device" | "claim" | "video" | "system" | "claim wherin" |
| --- | --- | --- | --- | --- |
| "internet" | "method" | "user" | "telephone" | "communication" |
| "device system" | "network" | "device method" | "information" | "system claim wherein" |
| "claim comprising step" | "remote" | "via" | "audio" | "system claim" |
| "server" | "monitoring" | "comprising step" | "wherein" | "method comprising" |
| "interface" | "wireless" | "wherein user" | "associated" | "personal" |

Figure 6: Top 30 features for "Wireless Network" keyword

The main goal for the litigation model is to differentiate between the litigated and non-litigated patents. The confusion matrix shows the performance of the model by comparing between the actual class (how a patent is originally labeled) and the predicted class (how the patent is predicted by the model). Positive and negative labels imply the litigation and non-litigation cases, respectively. We are interested in the probability that a patent is actually litigated when the predicted outcome is litigated, which is the precision. We also need to pay attention to the recall, which is the proportion of the patents that are predicted to be litigated among all litigated patents. Generally, there is an inverse relationship between these two values. Decreasing one value is compensated by a higher value of the other.

As it is not trivial to make a descriptive conclusion from considering both precision and recall, the F1-score representing the trade-off between the precision and recall is commonly used to compare the performance among different models. Different parameters such as the cut-off probability to define the predicted class and the regularization parameters as well as the hyper parameters used in the clustering approach are experimented to tune the model. The best values have been listed at the beginning of this section. For the litigation model, a comparison of the F1-score among different keywords and data options is depicted in Figure 7 while Figure 8 compares the precision and recall. Figure 8 illustrates no distinct pattern among data options, except the obvious inverse relationship between precision and recall. The data option of SEC without default gives the highest F1-score, which is consistent across the three keywords. Comparing among keywords, "Wireless Network" yields the best performance, followed by "Telecommunication" and "Advertising," respectively.

A comparison of the F1-score between the cluster with ensemble method and the pure classification approach for the SEC data without default across all 3 keywords is given in Table 4. The cluster with ensemble method for SEC without default data option outperforms the pure classification approach for the "Wireless Network" and "Telecommunication" keywords.

Figure 9 illustrates confusion matrices for "Wireless Network" with SEC without default option which perform best among all models. We observe a trade-off among 4 performance measures: true positive, false positive, false negative and true negative values with different sets of hyper parameters. The confusion matrices with different hyper parameters obtained from the pure classification approach are shown in Figure 10.

In the time-to-litigation model, the F1-scores obtained from the pure classification models are compared with those from the cluster with ensemble models for each time period as illustrated in Table 5. "Wireless Network" generally provides the best performance compared to the other keywords in almost all models and all data options except T < 1 class. The cluster with ensemble method performs better for the "Telecommunication" keyword while the pure classification method gives a better performance for the "Wireless Network" and "Advertising" keyword. Utilizing SEC data without default generally performs worse than other data options regardless of the choice of keywords. However, there is no obvious conclusion with respect to cluster with ensemble vs pure classification.

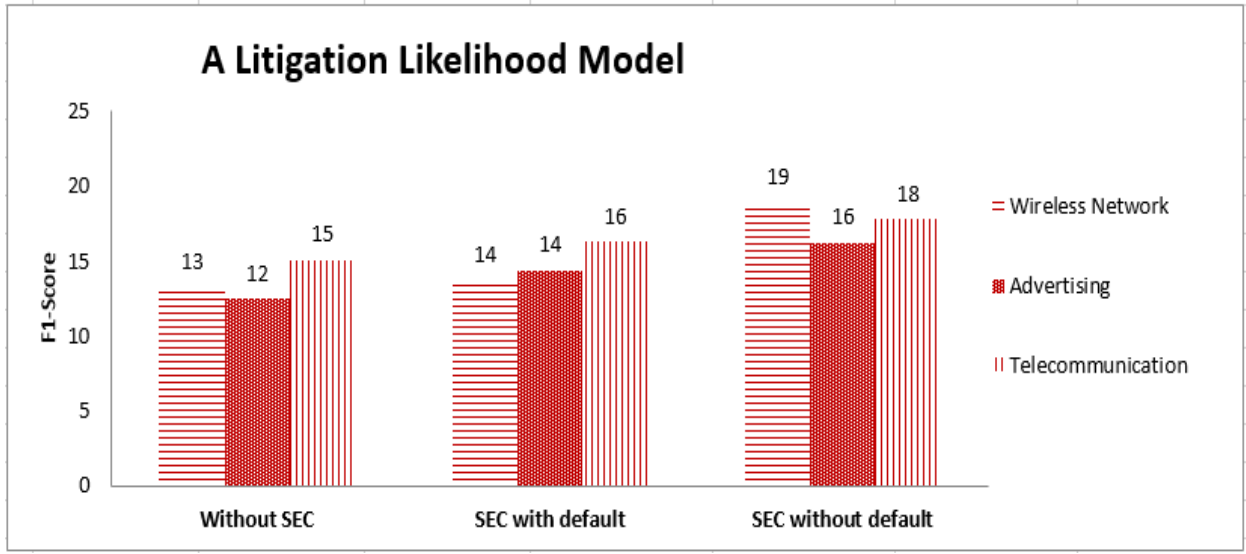

Figure 7: A comparison of F1-score as a percentage

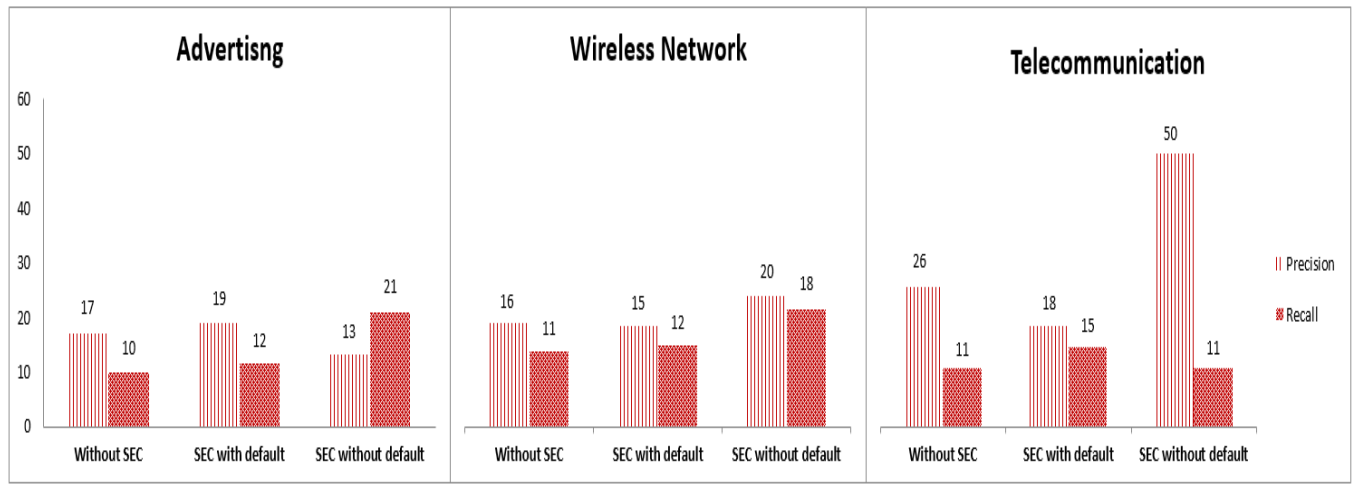

Figure 8: A comparison of precision and recall as a percentage

Table 4: A comparison of F1-score between cluster with ensemble and pure classification

|  | Pure classification | Cluster with ensemble |
|---|---|---|
| Wireless Network | 0.1554 | **0.1886** |
| Advertising | **0.1716** | 0.1623 |
| Telecommunication | 0.1711 | **0.1778** |

|  |  | Actual | |
|---|---|---|---|
|  |  | positive | negative |
| Predicted | positive | 24<br>28 | 92<br>113 |
| | negative | 132<br>128 | 7419<br>7398 |

Figure 9: The confusion matrix for "Wireless Network" keyword and SEC data without default with different hyper parameters for the cluster with ensemble method

|  |  | Actual | |
|---|---|---|---|
|  |  | positive | negative |
| Predicted | positive | 23<br>30 | 144<br>211 |
| | negative | 133<br>126 | 7367<br>7300 |

Figure 10: The confusion matrix for "Wireless Network" keyword and SEC data without default with different hyper parameters for the pure classification method

Table 5: A comparison of F1-score between two methods for time-to-litigation

| Keywords | T < 7 years class | | | | | |
|---|---|---|---|---|---|---|
| | Without SEC | | SEC with default | | SEC without default | |
| | pure classification | cluster with ensemble | pure classification | cluster with ensemble | pure classification | cluster with ensemble |
| Wireless Network | 0.8807 | 0.8535 | **0.8827** | 0.8339 | 0.8609 | 0.8078 |
| Advertising | **0.8629** | 0.8263 | 0.8611 | 0.8279 | 0.8604 | 0.8479 |
| Telecommunication | 0.8475 | **0.8492** | 0.8484 | 0.8448 | 0.7301 | 0.7182 |
| | T < 4 years class | | | | | |
| Keywords | Without SEC data | | SEC with default | | SEC without default | |
| | pure classification | cluster with ensemble | pure classification | cluster with ensemble | pure classification | cluster with ensemble |
| Wireless Network | 0.7708 | 0.7698 | **0.7737** | 0.7619 | 0.7269 | 0.6820 |
| Advertising | **0.7548** | 0.7454 | 0.7441 | 0.7439 | 0.7444 | 0.7511 |
| Telecommunication | 0.7382 | 0.7385 | 0.7442 | **0.7539** | 0.5098 | 0.5348 |
| | T < 1 year class | | | | | |
| Keywords | Without SEC | | SEC with default | | SEC without default | |
| | pure classification | cluster with ensemble | pure classification | cluster with ensemble | pure classification | cluster with ensemble |
| Wireless Network | 0.3773 | 0.3793 | 0.3758 | **0.3803** | 0.3428 | 0.3373 |
| Advertising | **0.4305** | 0.4072 | 0.4165 | 0.4068 | 0.3691 | 0.3976 |
| Telecommunication | 0.3271 | **0.3681** | 0.3426 | 0.3623 | 0.1824 | 0.1630 |

**Discussions**

In this paper, we develop a combined clustering and classification model to predict whether a patent is likely to be litigated and when it would happen. The best litigation model produces a 0.19 F1-score that is obtained from the "Wireless Network" keyword with SEC without default. Excluding SEC data gives worse performance than the other two models. This implies that financial information is beneficial.

Compared to Chien's model, which is the closest related work to ours, we use the data set that truly reflects the actual litigation rate. Not only significant imbalanced data causes the problem to be difficult, but also collecting a large number of correct data records is challenging. A litigation is found to be a very rare event with 1 to 2 percent. Our models with SEC without default yield approximately 0.13, 0.2 and 0.5 precision for three keywords (the probability of accurately predicting the litigation of a patent) under an acceptable value of recall. The improvement can be obviously recognized. With this value of precision the model incurs a relatively large number of false positives which is reflected by the recall value. The hyper parameters can be adjusted so that the desired balance level of precision and recall is achieved. This balance level depends mainly on the users' preference. For example, the users should pay more attention to increasing the precision value if their priority is to minimize the cost corresponding to missing the litigated patents.

An ability to correctly predict time to litigation of a patent by using the time-to-litigation model helps the users to save substantial resources. The model with SEC without default generally yields the worst performance regardless of keywords due to very limited data. Intuitively, the model corresponding to the top node gives a better F1-score than a bottom node. Among different classes, a high F1-score at a bottom node (node 6 in Figure 4) implies that the model performs well. Even though it is clear from the litigation model that enhancing the clustering with ensemble approach gives a better performance, this trend does not continue with the time-to-litigation model. We observe that the data in the time-to-litigation model is no longer strongly imbalanced and relatively small. This observation implies that the cluster with ensemble approach adapted from anomaly detection does not improve the performance in the time-to-litigate settings across the board.

The features related to the reference knowledge are important indicators for differentiating between litigated and non-litigated patents. Large numbers of references as well as a large number of litigated referenced patents imply a higher interest in that particular patent. The SEC data of each patent's assignee provides insight into the financial situation of the company owning the patent and improves the predictive power of the models.

**Conclusions and future work**

The proposed litigation and time-to-litigation models attempt to predict the litigation likelihood and when it would occur. The clustering with ensemble approach are implemented in order to provide reliable predictive models. The problem is very challenging due to the low rate of litigation as well as the difficulty in obtaining a complete data set. Hence, better models can possibly be achieved if more complete data sets are accessible. Future work can be done to improve the time-to-litigation model by considering a multi-class multi-label classification which takes hierarchical constraints into account and fit the global model [13]. An overall loss function is used to compare different models of each class separately. The loss function for a record , defined in [13] is a combination of penalty costs for misclassifying each node in the hierarchy tree depicted in Figure 4. The cost occurs at each class if the model misclassifies that particular class while its upper-classes in the hierarchy tree are correctly predicted.

# References

1. Powers, D.M.W., Evaluation: From precision, recall and F-measure to ROC, informedness, markedness and correlation. Journal of machine learning technologies, 2011. 2(1): p. 37-63.

2. Chien, C.V., Predicting patent litigation. Texas law review, 2011. 90(2): p. 283-329.

3. Petherbridge, L., On predicting patent litigation. Texas law review, 2012. 90(75): p. 75-86.

4. Kesan, J.P., Schwartz, D.L. and Sichelman, T.M., Paving the path to accurately predicting legal outcomes: A comment on professor Chien's predicting patent litigation. Texas law review, 2012. 90: p. 97-109.

5. Lanjouw, J.O. and Schankerman M., Characteristics of patent litigation: A window on competition. Rand journal of economics, 2001. 32(1): p. 129-151.

6. Lanjouw, J.O. and Schankerman M., Protecting intellectual property: Are small firms handicapped? Journal of law and economics, 2004. 47(1): p. 45-74.

7. Cremers, K., Determinants of patent litigation in Germany. ZEW - Center for European economic research discussion paper, 2004.

8. Cowart, T.W., Lirely, R. and Avery, S., Two methodologies for predicting patent litigation outcomes: Logistic regression versus classification trees. American business law journal, 2014. 51(4): p. 843-877.

9. Kashima, H., Hido, S., Tsuboi, Y., Tajima, A., Ueno, T., Shibata, N., Sakata, I. and Watanabe, T., Predicting modeling of patent quality by using text mining. Proceedings of the international association for management of technology, 2010.

10. Chandola, V., Banerjee, A. and Kumar, V., Anomaly detection: A survey. ACM Computing surveys (CSUR), 2009. 41(3): p. 1-58.

11. Nalbantov, G.I., Groenen, P.J.F. and Bioch, J.C., Nearest convex hull classification. Econometric instiute report, 2006.

12. Chawla, N.V., Bowyer, K.W., Hall, L.O. and Kegelmeyer, W.P., SMOTE: Synthetic minority over-sampling technique. Journal of artificial intelligence research, 2002. 16(1): p. 321-357.

13. Vens, C. and Struyf, J., Decision trees for hierarchical multi-label classification. Machine learning 2008. 73(2): p. 185-214.